\documentclass[11pt]{article}

\usepackage[final]{acl}

\usepackage{times}
\usepackage{latexsym}

\usepackage[T1]{fontenc}

\usepackage[utf8]{inputenc}

\usepackage{microtype}

\usepackage{inconsolata}

\usepackage{graphicx}

\usepackage{enumitem}
\usepackage{amsmath}
\usepackage{multirow}
\usepackage{booktabs}
\usepackage{color}
\usepackage{xcolor}
\usepackage{colortbl}
\usepackage{tcolorbox}
\usepackage{pifont}
\usepackage{amssymb}

\definecolor{rightgreen}{RGB}{62, 199, 97}
\definecolor{falsered}{RGB}{252, 86, 107}

\newcommand{\cmark}{\textcolor{rightgreen}{\ding{51}}}
\newcommand{\xmark}{\textcolor{falsered}{\ding{55}}}

\usepackage{silence}
\title{GeoArena: Evaluating Open-World Geographic Reasoning in Large Vision-Language Models}

\author{
    \textbf{Pengyue Jia\textsuperscript{1,2}\footnotemark[1]},
    \textbf{Yingyi Zhang\textsuperscript{1}\footnotemark[1]},
    \textbf{Xiangyu Zhao\textsuperscript{1}\footnotemark[2]},
    \textbf{Sharon Li\textsuperscript{2}}
\\
 \textsuperscript{1}Department of Data Science, City University of Hong Kong\\
 \textsuperscript{2}Department of Computer Sciences, University of Wisconsin-Madison
\\
 \small{
 \{jia.pengyue,yzhang6375-c\}@my.cityu.edu.hk, xianzhao@cityu.edu.hk, sharonli@cs.wisc.edu
 }
}

\begin{document}
\maketitle
\renewcommand{\thefootnote}{\fnsymbol{footnote}}
\footnotetext[1]{Equal contribution}
\footnotetext[2]{Corresponding author}
\renewcommand{\thefootnote}{\arabic{footnote}}

\begin{abstract}
    Geographic reasoning is a fundamental cognitive capability that requires models to infer plausible locations by synthesizing visual evidence with spatial world knowledge. Despite recent advances in large vision-language models (LVLMs), existing evaluation paradigms remain largely outcome-centric, relying on static datasets and predefined labels that are conceptually misaligned with open-world geographic inference. Such outcome-centric evaluations often focus exclusively on label matching, leaving the underlying linguistic reasoning chains as unexamined black boxes. In this work, we introduce \textbf{GeoArena}, a dynamic, human-preference-based evaluation framework for benchmarking open-world geographic reasoning. GeoArena reframes evaluation as a pairwise reasoning alignment task on in-the-wild images, where human judges compare model-generated explanations based on reasoning quality, evidence synthesis, and plausibility.
    We deploy GeoArena as a public platform and benchmark 17 frontier LVLMs using thousands of human judgments, which complements existing benchmarks and supports the development of geographically grounded, human-aligned AI systems. We further provide detailed analyses of model behavior, including reliability of human preferences and factors influencing judgments of geographic reasoning quality. %
    We open-source GeoArena\footnote{\url{https://github.com/Applied-Machine-Learning-Lab/ACL2026_GeoArena}} to foster future research.
\end{abstract}

\section{Introduction}

\begin{figure*}
    \centering
    \includegraphics[width=\textwidth]{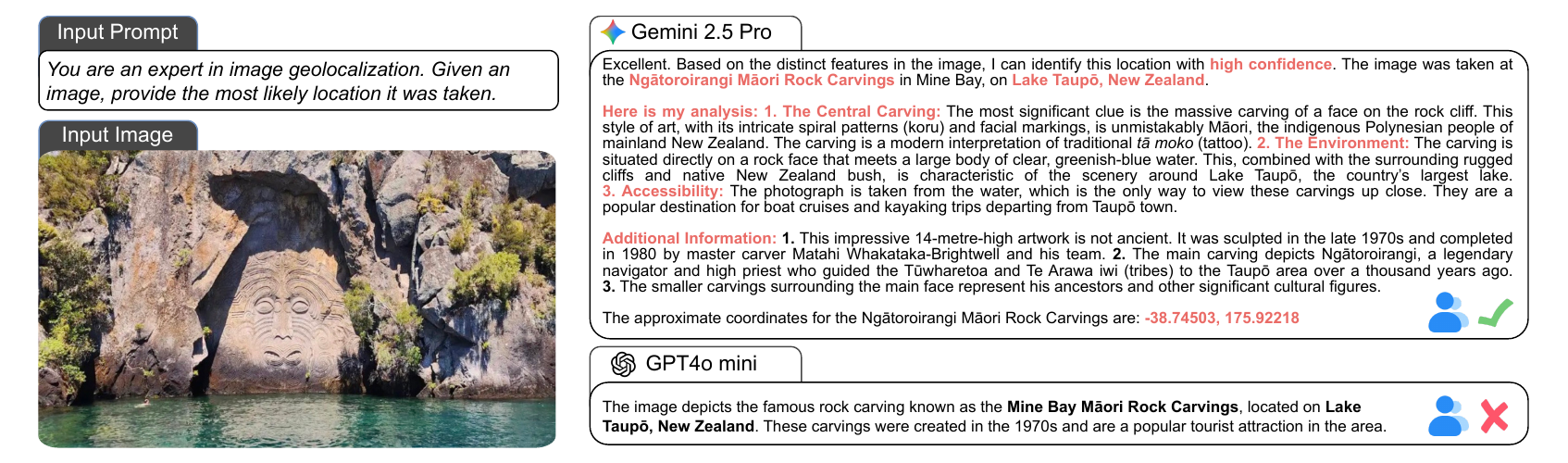}
    \caption{Example of geolocalization: identifying the Ngātoroirangi Māori Rock Carvings.}
    \label{fig:case_study}
\end{figure*}

Geographic reasoning--the ability to infer, contextualize, and explain where an observation could plausibly be situated in the world--is a fundamental capability for intelligent systems operating in the physical world~\cite{mai2022review}. It underlies applications ranging from navigation~\cite{venable2016improving} and disaster monitoring~\cite{li2025towards} to crime tracking~\cite{bamigbade2024computer}, and is commonly studied through image-based geolocalization as a canonical stress test~\cite{vo2017revisiting}, as illustrated in Figure~\ref{fig:case_study}. Crucially, geographic reasoning is inherently open-world: real-world images exhibit high variability in visual appearances across the globe~\cite{wilson2021visual}, which requires the integration of visual evidence with spatial, environmental, and cultural knowledge~\cite{li2025recognition}. As a result, geographic inference is under-determined by the input and intrinsically involves abductive reasoning and contextual judgment rather than deterministic prediction~\cite{wilson2021visual}.

Despite this open-world nature, existing evaluation paradigms for geographic reasoning remain largely outcome-centric, relying on static datasets with predefined labels and metrics such as coordinate distance or categorical accuracy~\cite{liu2024image,wang2024llmgeo,astruc2024openstreetview,li2025pixels,huang2025vlms,jay2025evaluating,yerramilli2025geochain}, as shown in Table~\ref{tab:benchmark_comparison}. While effective in controlled settings, this approach faces fundamental limitations in real-world scenarios: static benchmarks are increasingly vulnerable to data contamination from web-scale pretraining, outcome-level metrics collapse complex reasoning processes into a single label match, and authoritative ground truth is often unavailable or ambiguous for in-the-wild images. More fundamentally, label-based evaluation assumes a well-defined notion of correctness that does not hold when multiple hypotheses are plausible. In such settings, evaluating how models reason—how they select, justify, and contextualize evidence—provides a complementary and necessary signal that accuracy alone cannot capture.

\begin{table}
\centering
\caption{Comparison between existing benchmarks and GeoArena.}
\label{tab:benchmark_comparison}
\resizebox{\linewidth}{!}{
\begin{tabular}{lccccc} 
\toprule
Benchmarks         & Publication & Evaluation                            & \begin{tabular}[c]{@{}c@{}}Dynamic \\Datasets\end{tabular} & \begin{tabular}[c]{@{}c@{}}Reasoning \\Oriented\end{tabular} & \begin{tabular}[c]{@{}c@{}}Label\\Free\end{tabular}  \\ 
\midrule
OSV-5M             & CVPR'24    & GPS                                   &     \xmark &\xmark & \xmark                                                                        \\
LLMGeo             & CVPR'24    & Country                               &      \xmark &\xmark                                                     &               \xmark                                            \\
ETHAN              & Arxiv'24   & GPS                                   &      \xmark &\xmark                                                     &           \xmark                                                \\
Location-Inference & AAAI'25    & GPS, Country, City                    &       \xmark &\xmark                                                    &           \xmark                                                \\
GeoChain           & ACL'25     & GPS, Pass Score                       &       \xmark  &\cmark                                                   &           \xmark                                                \\
FairLocator        & Arxiv'25   & GPS, Street, City, Country, Continent &       \xmark  &\xmark                                                   &           \xmark                                                \\
IMAGEO-Bench       & Arxiv'25   & GPS, City, State, Country             &        \xmark &\xmark                                                  &            \xmark                                               \\ 
\midrule
GeoArena           & Ours       & User Preference                       &        \cmark  &\cmark                                                  &           \cmark                                               \\
\bottomrule
\end{tabular}}
\end{table}

To address these issues, we introduce \textbf{GeoArena}, a dynamic and realistic evaluation framework for benchmarking open-world geographic reasoning in LVLMs. GeoArena reframes evaluation as a pairwise reasoning alignment task: given an in-the-wild image and a geographic prompt, two anonymous models generate natural-language explanations of the image’s likely location, and human users vote for the response that better aligns with human geographic expectations in terms of reasoning quality, evidence synthesis, and plausibility. By design, GeoArena addresses the core limitations of existing evaluation paradigms: (1) it utilizes in-the-wild user contributions to mitigate {data contamination}; (2) it shifts the focus from outcome level to the assessment of {geographic reasoning chains} by evaluating the logical and linguistic quality of model explanations; (3) it employs a {human-preference-driven methodology}, reflecting real-world scenarios where precise metadata or ground truth label is unavailable. %
While human-preference-based evaluation is a standard protocol in other domains~\cite{chiang2024chatbot,jiang2024genai}, such a paradigm remains a significant gap in the geographic AI community. GeoArena bridges this gap and can facilitate developing geographic reasoning systems that are both accurate and fundamentally aligned with human-centric logic and real-world utility.

Using GeoArena, we benchmark 17 frontier LVLMs through thousands of human preference judgments collected on in-the-wild images. Our results reveal clear stratification among models, with frontier systems consistently outperforming smaller variants, and strong open-source families are closing the gap. Importantly, GeoArena produces stable rankings with high agreement between expert and crowd judgments, demonstrating that human preference signals provide a reliable and discriminative evaluation of geographic reasoning quality beyond outcome-level accuracy. Our key contributions can be summarized as follows:

\begin{enumerate}[leftmargin=*]
    \item We formalize the problem of open-world geographic reasoning evaluation and develop GeoArena, the first dynamic, preference-based framework to address long-standing issues of existing benchmarks.
    \item We conduct a comprehensive analysis of the collected user inputs and voting data to demonstrate the reliability and capabilities of GeoArena.
    \item We publicly release GeoArena to support research and development in related fields such as LVLM and geographic foundation models.
\end{enumerate}

\section{Related Work}

\paragraph{Benchmark of Geographic Reasoning.} Current research in geographic reasoning focuses on evaluating the cognitive logic of models across diverse spatial tasks. GEOBench-VLM~\cite{danish2025geobench} provides a framework for evaluating vision-language models on tasks such as scene understanding, object counting, and temporal analysis within geospatial contexts. Another study~\cite{huang2025evaluating} introduces a benchmark for geometry classification, topological relations, and direction estimation, which assesses core spatial logic using geometries encoded in GeoJSON format. Furthermore, MapEval~\cite{dihan2024mapeval} evaluates map-based reasoning through textual, visual, and API-based modes, identifying performance gaps in spatial inference tasks like distance and route planning. These geographic reasoning capabilities are frequently evaluated through image geolocalization, which serves as a stress test for synthesizing visual evidence with spatial world knowledge. Common evaluation datasets for geolocalization include IM2GPS~\cite{hays2008im2gps} and YFCC~\cite{thomee2016yfcc100m}. On the benchmarking side~\cite{yerramilli2025geochain}, LLMGeo~\cite{wang2024llmgeo} collects data from Google Street View to evaluate various models, while \citet{liu2024image} demonstrates that incorporating Chain-of-Thought (CoT) reasoning improves results on geolocalization tasks. \citet{jay2025evaluating} and FairLocator~\cite{huang2025vlms} provide generalized evaluation sets and focus on urban geolocalization biases, respectively. 
In contrast to these static methods, we propose GeoArena, the \emph{first dynamic and user-preference-based benchmark for geographic reasoning}. 
This approach provides a user-aligned platform for assessing geographic reasoning in real-world applications.

\paragraph{Worldwide Image Geolocalization.}
Worldwide image geolocalization~\cite{jia2025georanker,jia2026georouter} is an interdisciplinary task that bridges geography and computer science~\cite{zhang2026evoking,xu2025single,zhu2023difftraj,zhu2024controltraj}, involving GeoAI~\cite{janowicz2020geoai,han2025swarm,xu2016taxi,zhu2023synmob,kong2024mobile,zhu2024unitraj}, spatial data mining~\cite{wang2020deep,zhao2017exploring,zhao2022multi,zhang2023promptst,zhang2023autostl,cheng2025poi,zhao2017incorporating}, information retrieval~\cite{jia2024mill,jia2025bridging,zhang2023mlpst}, and multi-modal modeling~\cite{wang2023large,zhang2025notellm}. In recent years, thanks to the strong world knowledge and visual understanding capabilities of LVLMs, image geolocalization has made significant progress~\cite{vivanco2023geoclip,li2024georeasoner,haas2024pigeon,dou2024gaga,sarkar2024gomaa,astruc2024openstreetview,dufour2025around,li2025recognition}. 
Img2Loc~\cite{zhou2024img2loc} is the first to introduce LVLMs into image geolocalization, retrieving similar images' information and incorporating it as prompts into the LVLM input to utilize the world knowledge acquired during pretraining to predict the image's location. 
G3~\cite{jia2024g3} further improves upon Img2Loc by optimizing both the image retrieval and reasoning processes, enabling the model to obtain more accurate reference information and fully exploit the prediction potential of LVLMs.

\section{Problem Definition}

We formalize the evaluation of open-world geographic reasoning as a task of measuring the alignment between model-generated explanations and human geographic expectations under uncertainty. Let $\mathcal{I}$ be the space of in-the-wild images and $\mathcal{P}$ be the space of natural language prompts. For a pair $(I, P) \in \mathcal{I} \times \mathcal{P}$, a model $M$ produces a response $R$ in the space of linguistic reasoning $\mathcal{R}$.

We define the evaluation task as a mapping $\mathcal{A}: \mathcal{R} \times \mathcal{H} \to \mathbb{R}$, where $\mathcal{H}$ represents the latent space of Human Expectations. The reasoning capability of a model $M$ is evaluated as:\begin{equation}E_{\text{reasoning}}(M) = \mathbb{E}_{(I, P) \in \mathcal{I} \times \mathcal{P}} [\mathcal{A}(M(I, P), \mathcal{H})],\end{equation}where $\mathcal{A}$ measures the degree of alignment between the generated reasoning chain and the spatial logic expected by humans. This definition allows the evaluation to function in open-world scenarios where precise metadata is absent and enables a process-oriented assessment of how models synthesize visual evidence with spatial knowledge.

\section{GeoArena}

GeoArena is an interactive platform designed to evaluate the open-world geographic reasoning capabilities of various LVLMs. In this section, we provide a detailed description of GeoArena, including its system architecture and interface (Section~\ref{sec:live_interface}), data collection process (Section~\ref{sec:data_collection}), the models it evaluates (Section~\ref{sec:models}), and the ranking computation methods (Section~\ref{sec:ranking_methods}).

\subsection{System Architecture and Interface} \label{sec:live_interface}

\begin{figure*}
    \centering
    \includegraphics[width=\linewidth]{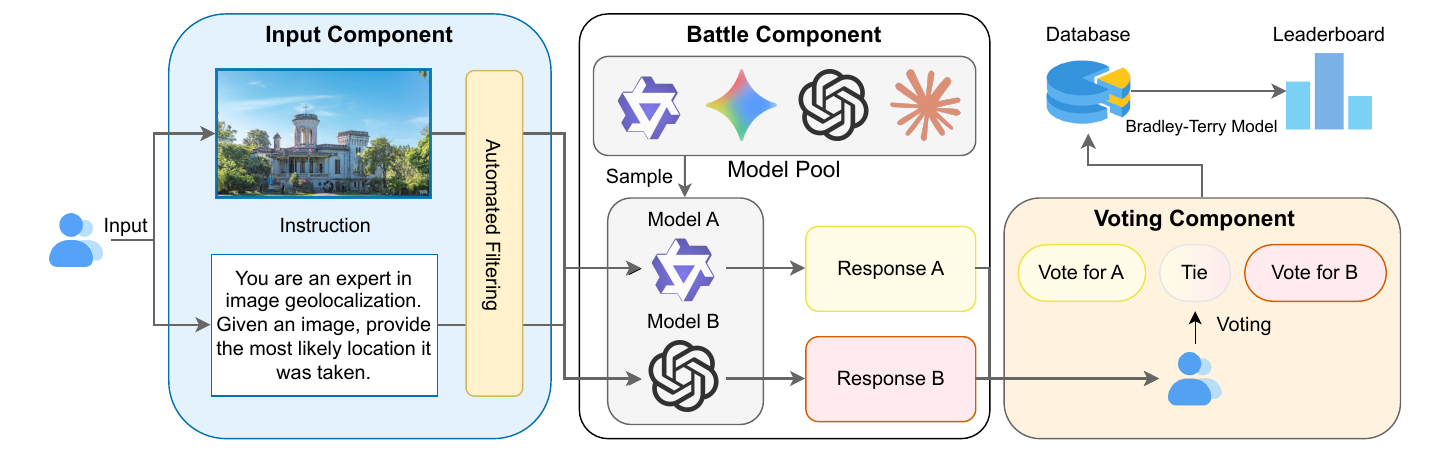}
    \caption{Overview of GeoArena.}
    \label{fig:overview}
\end{figure*}

GeoArena is designed to facilitate the collection of high-fidelity data on geographic reasoning. As illustrated in Figure~\ref{fig:overview}, the system architecture consists of a multi-stage pipeline integrated into a public-facing interface. This design is intended to ensure that every interaction contributes to a controlled and unbiased evaluation, transforming the user interface into a rigorous scientific instrument for observing model behavior. 

The \textbf{Input Component} serves as the primary gateway for data acquisition at the {input level}. Users contribute in-the-wild images and specify instructions. To maintain the thematic integrity of the platform and protect the leaderboard from noise, we implement an automated filtering mechanism as an architectural component powered by LLMs (more details in Appendix~\ref{sec:appendix_auto}). Every user prompt $P$ is processed by a classification function $\phi(P)$ to verify that the query is relevant to geographic reasoning. This layer proactively rejects irrelevant or malicious inputs, ensuring that the collected preferences reflect genuine spatial inference capabilities. For standard interactions, a default instruction is provided based on established protocols in the field~\cite{zhou2024img2loc, jia2024g3}. 

The \textbf{Battle Component} constitutes the second stage of the pipeline, addressing {process neglect} at the {output level}. Once a user input is validated, the system performs sampling to select two models, $M_A$ and $M_B$, from the participant pool. Each model is required to generate a response $R$. This stage is engineered to transform the black-box inference of LVLMs into an observable reasoning chain, allowing for a direct assessment of how models synthesize environmental cues and spatial knowledge. The responses are presented in a side-by-side, anonymized format to ensure that the evaluation is based solely on the logical consistency and linguistic quality of the geographic reasoning.

The \textbf{Voting Component} serves as the final {evaluation level}, where human preference data is captured. Users are instructed to judge the responses based on their alignment with human geographic expectations, particularly in terms of the logical groundedness and clarity of the reasoning. The interface provides three mutually exclusive options: ``vote for A'', ``vote for B'', and ``tie''. This preference-driven approach is a deliberate design choice to establish a reliable benchmarking signal in unconstrained scenarios where authoritative labels are absent. By aggregating these pairwise judgments, the platform establishes a leaderboard that reflects real-world utility and the ability of models to communicate spatial logic to human users. The true identities of the models are revealed only after a vote is submitted to prevent brand-related bias and maintain the scientific integrity of the experimental results.

\subsection{Data Collection} \label{sec:data_collection}

The data collection protocol of GeoArena is designed to capture the comprehensive reasoning trajectories and preference signals required for a systematic analysis of geographic logic. For every evaluation session, the system records the specific image $I$, the user prompt $P$, the anonymized model responses $R_A$ and $R_B$, and the resulting human preference signal $S \in \{1, 0, 0.5\}$. This multi-modal dataset ensures the traceability of the reasoning process and supports the reproducible computation of model rankings. All collected data are preserved in structured formats that facilitate downstream linguistic analysis and leaderboard updates. To preserve user privacy, we anonymize user inputs and apply filters to remove any potentially sensitive or inappropriate content. Please refer to Appendix~\ref{appendix:geoarena-1k} for more details.

\subsection{Participating Models} \label{sec:models}

To provide meaningful comparisons of geographic reasoning capabilities, GeoArena includes a diverse selection of both open-source and proprietary models. The selection includes representative Large Vision-Language Models (LVLMs) from multiple providers. For the GPT series~\cite{achiam2023gpt}, it evaluates GPT 4o, GPT 4o mini, GPT 4.1, GPT 4.1 mini, and GPT 4.1 nano. From the Gemini family~\cite{team2023gemini}, it includes Gemini 2.5 pro and Gemini 2.5 flash. The Claude series includes Claude Opus 4 and Claude Sonnet 4. It also evaluates Llama 4 maverick and Llama 4 scout~\cite{touvron2023llama}, as well as Gemma 3 models~\cite{team2025gemma} in sizes of 27B, 12B, and 4B. Additionally, the platform evaluates Qwen 2.5 VL models in sizes of 72B, 32B, and 7B~\cite{bai2025qwen2}. As shown in Appendix~\ref{appendix:models}, GeoArena currently benchmarks 17 models in total. This coverage allows users and researchers to evaluate model performance across different architectures, training paradigms, and geographic reasoning capabilities.

\subsection{Ranking Methods} \label{sec:ranking_methods}

\paragraph{Online Elo Rating.} The Elo rating system is a standardized approach to estimate the relative strength of different models based on pairwise comparisons. It provides an interpretable score that reflects the expected probability of one model outperforming another. Formally, given two models $M_i$ and $M_j$ with ratings $\gamma_i$ and $\gamma_j$, the expected probability that model $M_i$ will outperform model $M_j$ is defined as:
\begin{equation}
P(M_i \succ M_j) = \frac{1}{1 + 10^{(\gamma_j - \gamma_i)/\alpha}}
\end{equation}
where $\alpha$ is a scaling parameter that controls the spread of the probability function, typically set to 400. After observing the actual outcome $S_{ij}$, where $S_{ij} = 1$ if $M_i$ wins, $S_{ij} = 0.5$ for a tie, and $S_{ij} = 0$ if $M_i$ loses, the rating of model $M_i$ is updated as:$\gamma_i' = \gamma_i + K \cdot (S_{ij} - P(M_i \succ M_j)),$where $K$ is a learning rate that determines the adaptation speed. Although the Elo system is computationally efficient, it is sensitive to the order of matches--an effect that is undesirable for benchmarking static LVLMs. To ensure a stable and order-invariant ranking, we follow prior work~\cite{chiang2024chatbot} and apply the Bradley-Terry model~\cite{bradley1952rank} to estimate the final scores for the geographic reasoning task.

\paragraph{Bradley-Terry Model.} The Bradley-Terry (BT) model provides a principled method to estimate the relative strength of competing models through the maximum likelihood of all observed outcomes. In this framework, each model $M_i$ is assigned a latent strength parameter $\gamma_i$. The probability $P(M_i \succ M_j)$ remains consistent with the Elo formulation. The BT model estimates the parameters $\gamma_i$ by maximizing the likelihood of all recorded pairwise comparisons, accounting for repeated trials through a weighting term $W_{ij}$. The likelihood function is defined as:
\begin{equation}
\mathcal{L}(\mathbf{\Gamma}) = \sum_{i,j \in N, i \neq j} W_{ij} \log \left( \frac{1}{1 + 10^{(\gamma_j - \gamma_i)/\alpha}} \right)
\end{equation}
To compute the final ratings, we apply a linear transformation to align the scores with the standard Elo scale. After fitting the BT model via logistic regression, the estimated parameters $\hat{\gamma}_i$ are transformed as:
$\text{rating}_i = \text{scale} \cdot \hat{\gamma}_i + \gamma_{\text{base}},$
where \text{scale} is set to 400 and $\gamma_{\text{base}}$ is set to 1000. This transformation preserves the relative ranking while ensuring the scores are consistent with established benchmarking conventions.

\paragraph{Confidence Interval.} To ensure that the ranking results are not dependent on a specific sample of comparisons, we estimate confidence intervals (CIs) for the ratings. We adopt a bootstrap procedure~\cite{chiang2024chatbot} that repeatedly resamples the battle outcomes and re-computes the estimates. This approach allows us to quantify the variability in model rankings and provides statistically grounded intervals. The inclusion of confidence intervals is essential to distinguish between meaningful performance differences and those that arise from sampling noise. As a result, the reported rankings offer stronger evidence of the relative strengths of different LVLMs in geographic reasoning.

\section{Benchmarks and Results Analysis}

\subsection{Arena Leaderboard}

\begin{table}
\centering
\caption{GeoArena Leaderboard.}
\label{tab:leaderboard}
\resizebox{\linewidth}{!}{
\begin{tabular}{lccc} 
\toprule
Model                        & ELO Rating & 95\% CI lower & 95\% CI upper  \\ 
\midrule
Gemini 2.5 pro       & 1319.7     & 974.8         & 1443.8         \\
Gemini 2.5 flash     & 1206.5     & 1062.2        & 1330.6         \\
 Qwen 2.5 VL 72B & 1094.5     & 982.6         & 1181.9         \\
 Gemma 3 12B       & 1086.5     & 1002.6        & 1186.4         \\
 Gemma 3 27B       & 1065.5     & 959.3         & 1159.8         \\
 GPT 4.1 mini         & 1059.8     & 970.0         & 1161.4         \\
 Llama 4 maverick & 1046.6     & 944.6         & 1115.3         \\
 Qwen 2.5 VL 32B & 1044.8     & 964.9         & 1119.0         \\
 GPT 4.1              & 1044.8     & 964.9         & 1119.0         \\
 Claude Opus 4     & 1042.3     & 933.8         & 1130.0         \\
 Gemma 3 4B        & 1027.3     & 936.3         & 1102.0         \\
 Claude Sonnet 4   & 1019.9     & 921.3         & 1113.8         \\
 GPT 4o               & 1000.0     & 1000.0        & 1000.0         \\
 Llama 4 scout    & 984.2      & 876.0         & 1077.1         \\
 Qwen 2.5 VL 7B & 950.9      & 868.4         & 1056.2         \\
 GPT 4.1 nano         & 917.9      & 819.1         & 1015.5         \\
 GPT 4o mini          & 871.6      & 715.2         & 1114.7         \\
\bottomrule
\end{tabular}}
\end{table}

Table~\ref{tab:leaderboard} presents the GeoArena leaderboard, representing the geographic reasoning capability of 17 frontier models. The reported 95\% confidence intervals are estimated via bootstrap resampling over 100 rounds to account for rating variability across different voting subsets. To maintain the scientific integrity of the experimental results, the input data is processed through the automated quality control layer and manual filtering protocols established in Section~\ref{sec:live_interface}. Several observations can be identified from the results. First, Gemini models demonstrate the highest performance, with Gemini 2.5 pro and Gemini 2.5 flash outperforming all other evaluated systems. This suggest that large-scale, production-level pre-training provides a clear advantage in synthesizing visual cues for geographic reasoning. Second, open-source families such as Qwen 2.5 and Gemma 3 achieve competitive rankings. For instance, Qwen 2.5 VL 72B outperforms Gemma 3 12B and performs comparably to the GPT 4.1 series, indicating that open-source systems are rapidly reducing the gap with proprietary frontier models. Third, several models, including Llama 4 maverick, GPT 4.1, and Claude Opus 4, cluster within the Elo 1040 to 1050 range with significant overlap in confidence intervals, suggesting that their performance differences are not statistically significant. Fourth, smaller variants such as GPT 4.1 nano and GPT 4o mini exhibit a clear decrease in performance. This confirms the difficulty of geographic reasoning, where limited model capacity hinders generalization across diverse global environments. Finally, the wide rating distribution validates that GeoArena is an effective experimental apparatus for distinguishing between frontier systems and lightweight baselines. This differentiation is essential for advancing research on human-aligned geographic reasoning in LVLMs.

\subsection{Battle Data Analysis}
\begin{figure*}[h]
    \centering
    \includegraphics[width=\linewidth]{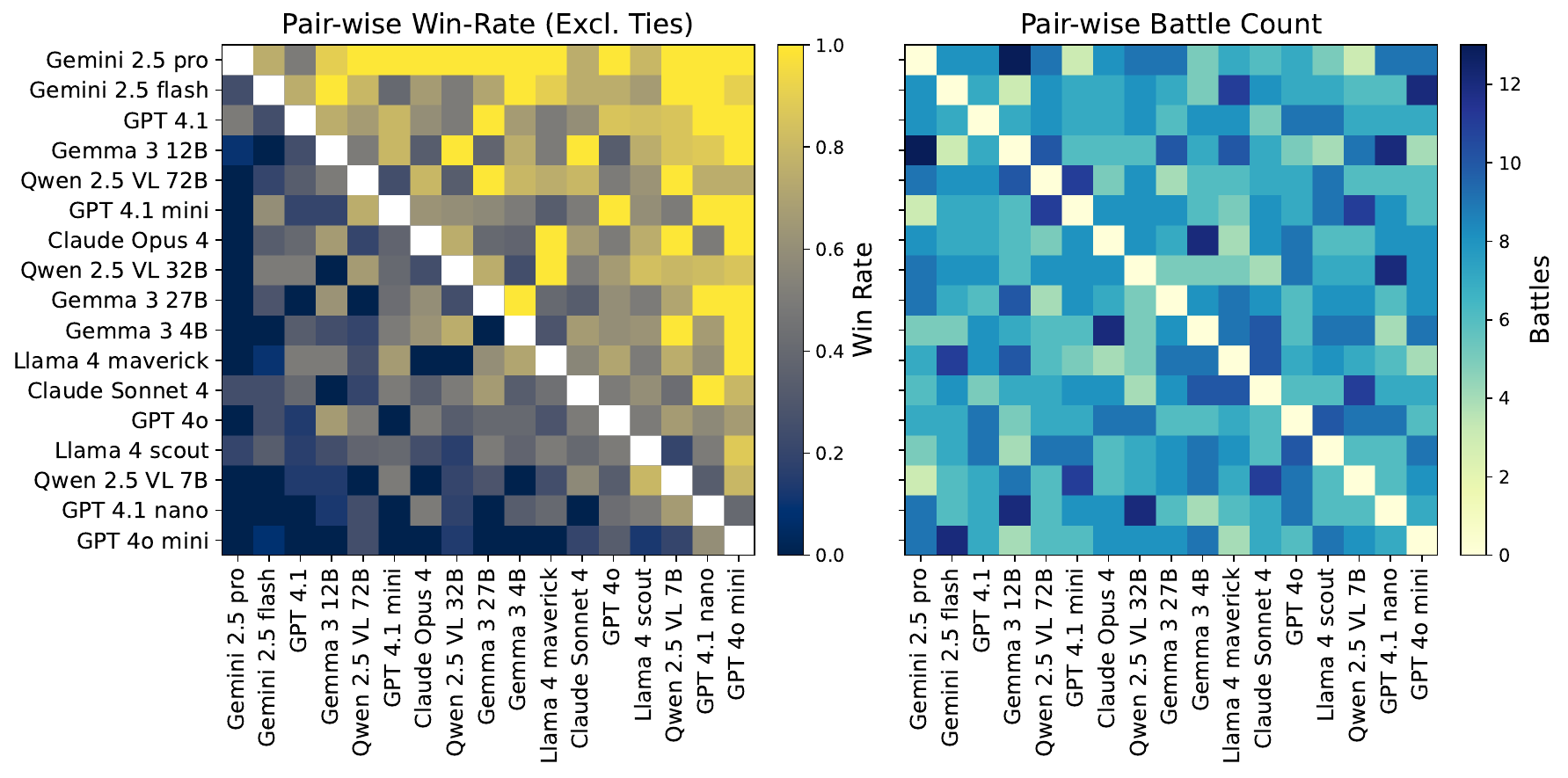}
    \caption{Pair-wise Performance Comparison of Models (Win-Rate and Battle Count). }
    \label{fig:pairwise_heatmap}
\end{figure*}

To provide a comprehensive view of geographic reasoning capabilities, we conduct a pairwise analysis of model interactions, reporting both win rates and battle counts. Figure~\ref{fig:pairwise_heatmap} illustrates the pairwise comparison across models, where the left panel displays head-to-head win rates and the right panel displays the corresponding battle counts. Models are ordered by their average win rate, which makes the hierarchy of geographic reasoning capabilities across the leaderboard interpretable.

Several findings are identified from this analysis. \textbf{(1) Frontier systems demonstrate consistent dominance in geographic reasoning.} Gemini 2.5 pro, Gemini 2.5 flash, and GPT 4.1 occupy the top rows, maintaining win rates close to or above 0.7 against nearly all competitors. This persistent advantage suggests that both model capacity and advanced alignment procedures contribute to the robustness of their spatial world knowledge and the logical consistency of their reasoning chains. \textbf{(2) Mid-scale models exhibit transitional behavior.} Models such as Gemma 3 12B, Qwen 2.5 VL 72B, and GPT 4.1 mini occupy the middle tier. They achieve favorable outcomes against smaller variants but exhibit substantial performance gaps when challenged by the frontier tier. This suggests a stratification that correlates with the ability to synthesize visual evidence into coherent geographic explanations. \textbf{(3) Lower-capacity systems show systematic reasoning deficits.} Models including Gemma 3 4B, Qwen 2.5 VL 7B, GPT 4.1 nano, and GPT 4o mini cluster near the bottom of the heatmap, with win rates typically below 0.3 against larger peers. These deficits are consistent across model families, reflecting limited parameter budgets that restrict the internal representation of global geographic contexts. \textbf{(4) Predictable scaling patterns in geographic logic.} Within model families, the quality of geographic reasoning scales predictably with parameter count. For example, the Qwen 2.5 VL series shows clear improvements when moving from 7B to 72B parameters. These trends suggest that scaling determines the consistency and depth of the reasoning process when interpreting environment-specific cues.

\subsection{Reliability Analysis of Voting}

\begin{table}
\centering
\caption{Agreement Analysis between Expert and Crowd.}
\label{tab:appendix_agreement}
\resizebox{\linewidth}{!}{
\begin{tabular}{c|cccc} 
\toprule
Expert \textbackslash{} Crowd & Left Win & Tie    & Right Win & Agreement Rate  \\ 
\midrule
Left Win                      & 30       & 3      & 3         & 83.3\%          \\
Tie                           & 5        & 21     & 6         & 65.6\%          \\
Right Win                     & 2        & 3      & 27        & 84.4\%          \\
Agreement Rate                & 81.1\%   & 77.8\% & 75.0\%    & 78.0\%          \\
\bottomrule
\end{tabular}}
\end{table}

To validate the reliability of the preference data, we randomly sample 100 instances from the dataset for expert review. In this procedure, an expert is presented with an image, a geographic reasoning prompt, and two anonymized model responses. The expert is instructed to evaluate which response demonstrates superior geographic logic and evidence synthesis. The expert is permitted to use external resources, such as search engines and map services, to verify the factual accuracy of the spatial claims. Each expert evaluation requires approximately 3 to 5 minutes to ensure a rigorous assessment of the reasoning process. Table~\ref{tab:appendix_agreement} illustrates the distribution of preferences between expert and crowd annotations for the sampled instances. We identify a consistently high agreement rate between expert and crowd judgments, ranging from 75\% to 85\%, with an average agreement of \textbf{78}\%. According to established studies~\cite{chiang2024chatbot}, this level of consensus represents strong agreement. These results demonstrate that the preference signals captured by GeoArena are reliable and accurately reflect geographic reasoning quality.

\subsection{Alignment Study}
\begin{table}[t]
    \centering
    \caption{Alignment accuracy of LLMs with human judgments on sampled response pairs.}
    \vspace{0pt}
    \label{tab:llm_alignment}
    \resizebox{0.85\linewidth}{!}{
    \begin{tabular}{lcc}
    \toprule
    Model     &  Gemini 2.5 pro & Qwen 2.5 VL 72B \\
    \midrule
    Accuracy     & 0.6579 & 0.4667 \\
    \bottomrule
    \end{tabular}}
\end{table}

\begin{figure*}[!t]
    \centering
    \includegraphics[width=0.99\linewidth]{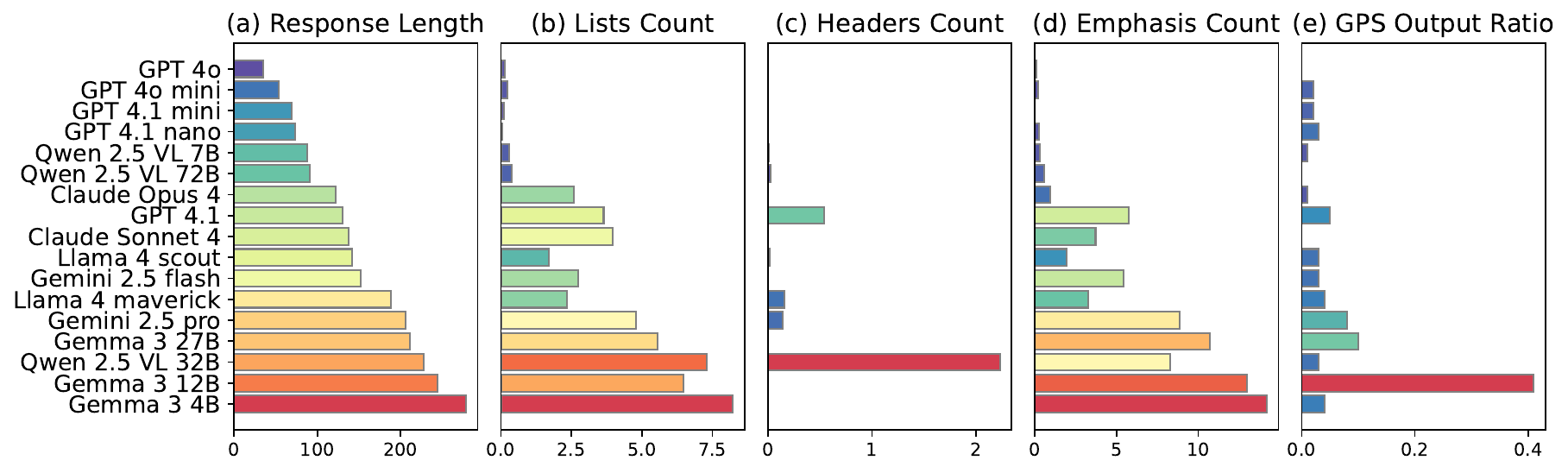}
    \caption{Distribution of Style Features in Model Outputs. }
    \label{fig:style_distribution}
\end{figure*}

To further examine whether LVLMs can act as reliable evaluators for geographic reasoning, we conduct an alignment study that compares model preferences with human judgments. Specifically, we randomly sample 100 response pairs from the dataset. For each pair, an LVLM is required to determine which response demonstrates superior geographic logic, evidence synthesis, and clarity of expression. The model is restricted to a single output label: win, tie, or loss. The prompt template for this task is provided in Appendix~\ref{appendix:llm_prompt}. We then calculate the concordance between the model's automated judgment and the human-provided preference labels.

Table~\ref{tab:llm_alignment} reports the alignment results for Gemini 2.5 pro and Qwen 2.5 VL 72B, representing the top proprietary and open-source models on the GeoArena leaderboard. The results show that Gemini 2.5 pro achieves a higher agreement rate (65.79\%) with human evaluations compared to Qwen 2.5 VL 72B (46.67\%). This suggests that Gemini 2.5 pro demonstrates stronger alignment with human geographic expectations when assessing reasoning chains.

These findings indicate that while current LVLMs can approximate human preferences to a certain extent, significant gaps remain. This observation supports the requirement for developing more faithful and robust automated evaluators for geographic reasoning and other multimodal tasks.

\subsection{Case Study}
To illustrate our framework, we present a case study using an image of the Ngātoroirangi Māori Rock Carvings at Mine Bay on Lake Taupō, New Zealand. As shown in Figure~\ref{fig:case_study}, different models exhibit varying levels of reasoning depth and factual accuracy. Gemini 2.5 pro produces a comprehensive analysis, identifying salient visual features such as the Māori face carving, surrounding cliffs, and water-based accessibility, while also providing historical and cultural context (e.g., the carving's creation in 1980 by Matahi Whakataka-Brightwell). In contrast, GPT 4o mini generates only a brief description, lacking explicit reasoning and omitting cultural details. This comparison underscores the importance of reasoning quality and contextual grounding in geolocalization tasks, showing that structured analyses align more closely with human preferences and task requirements. 
We also give hard cases analysis and more case studies in Appendix~\ref{appendix:hard_cases} and Appendix~\ref{appendix:additional_cases}. 

\subsection{Preference Analysis}

To identify which linguistic characteristics of geographic reasoning chains drive human preference, we extend the standard Bradley-Terry regression framework by incorporating style-related features as confounding variables following previous work~\cite{chiang2024chatbot,stylearena2024,dubois2024length}. By including these features in the regression, we can separate the effect of linguistic style from the intrinsic reasoning ability of the model. The style coefficients ($\beta$) are estimated via logistic regression, where normalized style features are included alongside model indicators in an extended design matrix. The resulting coefficients quantify the degree to which specific stylistic traits influence human judgments. 

In this study, we consider five different features: response length (measured by word count), list count (reflecting structured reasoning steps), header count, emphasis count (including bold and italic items), and GPS output ratio (the proportion of responses providing precise spatial claims). Figure~\ref{fig:style_distribution} illustrates the distribution of these features across different models, showing variation in the linguistic expression of geographic logic. Our analysis identifies findings consistent with prior studies in other domains~\cite{chiang2024chatbot,steyvers2024calibration,stylearena2024}. Specifically, response length exhibits a strong positive correlation with human preference ($\beta_{\text{length}}=0.526$), suggesting that longer explanations are perceived as providing more exhaustive geographic evidence. In addition, both list count ($\beta_{\text{list}}=0.095$) and GPS output ratio ($\beta_{\text{GPS}}=0.06$) are positively correlated with preference. A higher number of lists often reflects more explicit reasoning steps, while the presence of GPS coordinates provides a concrete anchor for the spatial reasoning chain. However, header count ($\beta_{\text{header}}=-0.153$) and emphasis count ($\beta_{\text{emphasis}}=-0.117$) do not show positive associations with human preference. It is possible that excessive structural markers or textual emphasis are perceived as superficial formatting rather than substantive geographic evidence, and thus do not contribute to the perceived informativeness of the reasoning process. The style-adjusted leaderboard is provided in Appendix~\ref{app:style-adjusted}.

\section{Conclusion}

In this work, we introduce \textbf{GeoArena}, a formal evaluation regime for benchmarking open-world geographic reasoning in LVLMs. By leveraging in-the-wild user contributions and pairwise human judgments, GeoArena addresses the critical failures of existing static benchmarks, including data contamination, reasoning process neglect, and ground-truth scarcity. Our framework decouples the assessment of logical reasoning from the hard requirement of exact coordinate labels, allowing for a more nuanced understanding of how models synthesize visual evidence with spatial knowledge. Through the implementation of a robust Bradley-Terry model, we establish a reliable leaderboard that reflects human geographic expectations.
GeoArena is open-sourced to foster future research. It is anticipated that GeoArena will support the related field in developing geographic reasoning systems that are both logically consistent and aligned with real-world utility.

\clearpage

\section*{Limitations}
While GeoArena establishes a formal regime for evaluating geographic reasoning, several limitations remain. First, although the platform is open to global participation, the geographic distribution of user-submitted images and votes may exhibit inherent bias based on the current user base. It is anticipated that as the platform attracts a broader demographic over time, this bias will naturally decrease. %
Moreover, to prioritize user privacy and encourage participation, the platform does not currently track unique user identifiers. This makes it difficult to quantify the impact of individual user bias on the overall leaderboard. Exploring privacy-preserving tracking methods, such as hashed identifiers, represents a potential solution for future development. Despite these challenges, GeoArena provides a unique and essential signal for human-aligned geographic reasoning that traditional static metrics cannot capture.

\section*{Ethical considerations}

The development and deployment of GeoArena as a formal evaluation regime for geographic reasoning adhere to rigorous ethical standards. We prioritize user privacy and data protection throughout the data acquisition pipeline. The platform does not collect or store personally identifiable information, and users are not required to submit metadata or coordinates tied to private locations. All contributed images and preference votes undergo an anonymization process and are managed in compliance with ethical data management practices. Our human evaluation methodology is restricted to pairwise preference voting and does not involve the collection of sensitive demographic or personal data. No financial compensation or targeted recruitment was involved in the data collection process. These considerations ensure that GeoArena functions as a scientific instrument for the geographic AI community. We use Large Language Models to polish the writing in this work.

\section*{Acknowledgments}
Pengyue Jia, Yingyi Zhang, and Xiangyu Zhao are supported by Hong Kong Research Grants Council (Research Impact Fund No.R1015-23, Collaborative Research Fund No.C1043-24GF, General Research Fund No. 11218325), and the Institute of Digital Medicine of City University of Hong Kong (No.9229503).

\bibliography{8Reference}

\clearpage
\appendix
\section{Appendix}

\subsection{Verifying the Feasibility of Automatic Prompt Filtering} \label{sec:appendix_auto}

To maintain a reliable leaderboard, it is essential to ensure that user inputs are relevant to the image's geographic reasoning. To examine whether LLMs can replace manual filtering, we conduct an experiment to determine whether LLMs can identify when a user prompt requests geolocating an image.

For this study, we construct a binary classification task. We randomly select 100 prompts from our voting data and assign them the label True, indicating that they ask about image geolocation. In parallel, we sample 100 prompts from the Chatbot Arena dataset\footnote{\url{https://huggingface.co/datasets/lmsys/chatbot_arena_conversations}}, which contains general-purpose prompts, and labeled them as False. Each model is given a simple instruction that defines image geolocation, specifies the expected JSON output, and directs the model to respond only with a True or False label. The instruction is given as follows:

\begin{tcolorbox}[colback=gray!5!white,colframe=gray!75!black]

You are a prompt classifier. Analyze the provided user prompt and determine if it is asking about image geolocalization.

Image geolocalization refers to determining or estimating the geographic location (e.g., city, country, landmark) where an image was taken based on its visual content.

Return ONLY a JSON object with one key: ``$\text{is\_geo}$''.
The value must be ``true'' if the prompt is inquiring about geolocalizing an image (e.g., ``Where was this photo taken?'' or instructions for an expert in image geolocalization), or ``false'' otherwise.
If uncertain, default to ``false''.

Output format (no extra words):
{{``$\text{is\_geo}$'': ``true''|``false''}}

User prompt: {$\text{user\_prompt}$}

\end{tcolorbox}

We evaluate three models: Gemini 2.0 flash, GPT 3.5 turbo, and GPT 4.1 mini. All three models achieve 100\% accuracy on this task. The high accuracy is mainly due to two factors. First, most users ask questions through the default prompt provided by GeoArena, which reflects a stable phrasing pattern. Second, prompts that request geographic reasoning usually contain explicit references to places, images, or location inference, which makes them easy for the models to detect. These observations show that modern language models can serve as reliable automatic filters for user inputs. Such a mechanism would allow the leaderboard to remain focused on geographic reasoning queries while reducing the need for manual inspection.

\subsection{Dataset Characteristics and Composition} \label{appendix:geoarena-1k}

\begin{figure}[h]
    \centering
    \includegraphics[width=0.9\linewidth]{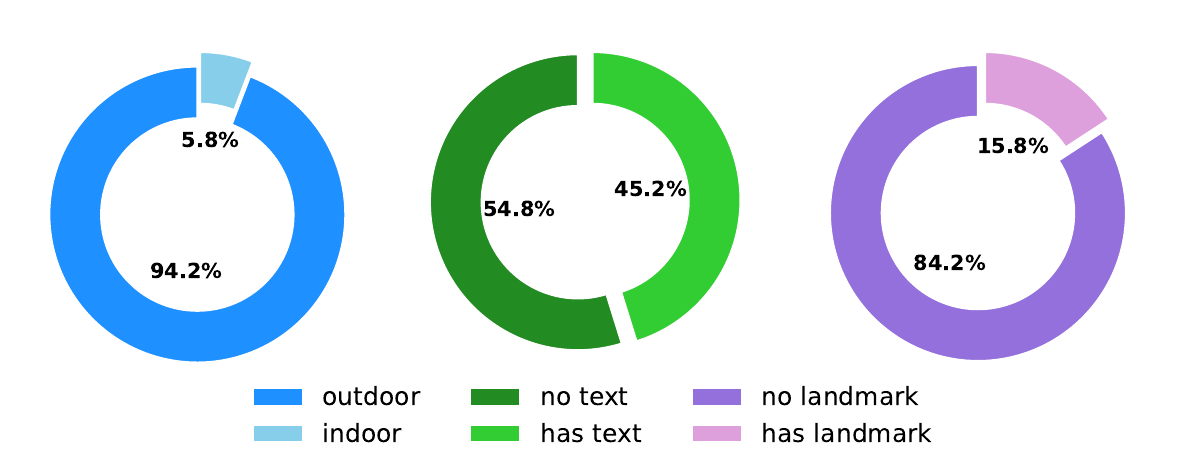}
    \caption{Composition of Image Features in Dataset}
    \label{fig:geoarena-1k-composition}
\end{figure}

To further explore the characteristics of the dataset, we employ GPT 4o to annotate the collected images, focusing on three key aspects:
\begin{enumerate}[leftmargin=*]
    \item Scene Type: whether the image depicts an indoor or outdoor setting.
    \item Text Presence: whether the image contains prominent, recognizable text.
    \item Landmark Presence: whether the image features a landmark, such as a historical site or natural icon.
\end{enumerate}

The corresponding results are presented in Figure~\ref{fig:geoarena-1k-composition}. The figure comprises three doughnut charts, each illustrating the distribution of one of the annotated attributes across the dataset: 
\begin{enumerate}[leftmargin=*]
    \item The first doughnut chart indicates that 94.2\% of images are classified as outdoor scenes, with only 5.8\% representing indoor environments. This pronounced skew toward outdoor imagery aligns with the global scope of GeoArena, where user-submitted images are likely dominated by exterior scenes captured in diverse geographic contexts.
    \item Text Presence: The second doughnut chart reveals a more balanced distribution, with 54.8\% of images lacking recognizable text ("no text") and 45.2\% containing text ("has text"). This near-equitable split underscores the dataset's richness, incorporating both text-free natural scenes and images with textual elements such as signs or labels. This variability is particularly valuable for assessing LVLM capabilities in multi-modal reasoning, where text recognition can enhance location prediction accuracy.
    \item Landmark Presence: The third doughnut chart shows that 84.2\% of images do not contain landmarks ("no landmark"), while 15.8\% do ("has landmark"). The low prevalence of landmarks reflects the dataset's emphasis on general geographic scenes rather than iconic or tourist-heavy locations, offering a broad representation of natural and urban environments worldwide. This distribution highlights the dataset's potential to test LVLM generalization across less distinctive locales, a challenging yet realistic scenario for global geolocalization. Overall, these distributions reveal the dataset's heterogeneity, making it a robust resource for benchmarking LVLM performance under real-world conditions.
\end{enumerate}

\subsection{Models in GeoArena} \label{appendix:models}

\begin{table*}[!ht]
\centering
\caption{Large-scale models benchmarked in \textit{GeoArena}.  Prices are USD / million tokens (input/output) and USD / thousand (image).}
\label{tab:models}
\vspace{5pt}
\resizebox{0.9\linewidth}{!}{
\begin{tabular}{lcccc} 
\toprule
\textbf{Model}    & \textbf{Company} & \textbf{Params} & \textbf{Openness} & \textbf{API Price (input / output / image)}  \\ 
\midrule
GPT 4o            & OpenAI           & Unknown               & Proprietary               & \$2.50 / \$10.00 / \$3.61                   \\
GPT 4o mini        & OpenAI           & Unknown               & Proprietary               & \$0.15 / \$0.60 / \$0.22                    \\
GPT 4.1           & OpenAI           & Unknown               & Proprietary               & \$2.00 / \$8.00 / -                          \\
GPT 4.1 mini      & OpenAI          & Unknown               & Proprietary               & \$0.40 / \$1.60 / -                          \\
GPT 4.1 nano      & OpenAI           & Unknown               & Proprietary               & \$0.10 / \$0.40 / -                          \\
Gemini 2.5 flash  & Google DeepMind  & Unknown               & Proprietary               & \$0.15 / \$0.60 / \$0.62                    \\
Gemini 2.5 pro    & Google DeepMind  & Unknown               & Proprietary               & \$1.25 / \$10.00 / \$5.16                    \\
Claude Sonnet 4   & Anthropic        & Unknown               & Proprietary               & \$3.00 / \$15.00 / \$4.80                    \\
Claude Opus 4     & Anthropic        & Unknown               & Proprietary               & \$15.00 / \$75.00 / \$24.00                  \\
Llama 4 maverick  & Meta             & 17B/402B             & Open-source                 & \$0.15 / \$0.60 / \$0.67                    \\
Llama 4 scout     & Meta              & 17B/109B             & Open-source                 & \$0.08 / \$0.30 / -                          \\
Gemma 3 27B       & Google           & 27B             & Open-source                 & \$0.10 / \$0.20 / \$0.03                    \\
Gemma 3 12B       & Google            & 12B             & Open-source                 & \$0.15 / \$0.10 / -                          \\
Gemma 3 4B        & Google            & 4B              & Open-source                 & \$0.02 / \$0.04 / -                          \\
Qwen 2.5 VL 72B   & Alibaba           & 72B             & Open-source                 & \$0.25 / \$0.75 / -                          \\
Qwen 2.5 VL 32B   & Alibaba          & 32B             & Open-source                 & \$0.90 / \$0.90 / -                          \\
Qwen 2.5 VL 7B    & Alibaba          & 7B              & Open-source                 & \$0.20 / \$0.20 / -                          \\
\bottomrule
\end{tabular}}
\end{table*}

GeoArena benchmarks 17 models in total, please refer to Table~\ref{tab:models} for details.

\subsection{LVLM Alignment Evaluation Prompt} \label{appendix:llm_prompt}

The prompt template used for LVLM alignment evaluation is as follows:

\begin{tcolorbox}[colback=gray!5!white,colframe=gray!75!black]

You are an expert evaluator in image geolocation tasks.  
I will give you two model responses to the same geolocation prompt.  

\textbf{Here is the prompt:}  

- Prompt: \texttt{\{sample['prompt']\}}  

- Image: \texttt{\{sample['image']\}}

\textbf{Response A:}  
\texttt{\{sample['response A']\}}  

\textbf{Response B:}  
\texttt{\{sample['response B']\}}  

\textbf{Your task is to decide which response is better based on:}  

1. Accuracy of the predicted location  

2. Strength of reasoning and evidence

3. Clarity and specificity  

\textbf{Output only one word:}  

- ``win'' if Response A is better  

- ``loss'' if Response B is better  

- ``tie'' if both are equally good  

\end{tcolorbox}

We first prompt the LVLM to act as an expert evaluator for the geographic reasoning task. For each sample, the model is provided with the prompt and the associated images, along with the responses from two candidate models. It is then asked to determine which response is better, considering three dimensions: accuracy, reasoning, and clarity and specificity.

\subsection{Difficult Queries Analysis} \label{appendix:hard_cases}

\begin{figure}
    \centering
    \includegraphics[width=\linewidth]{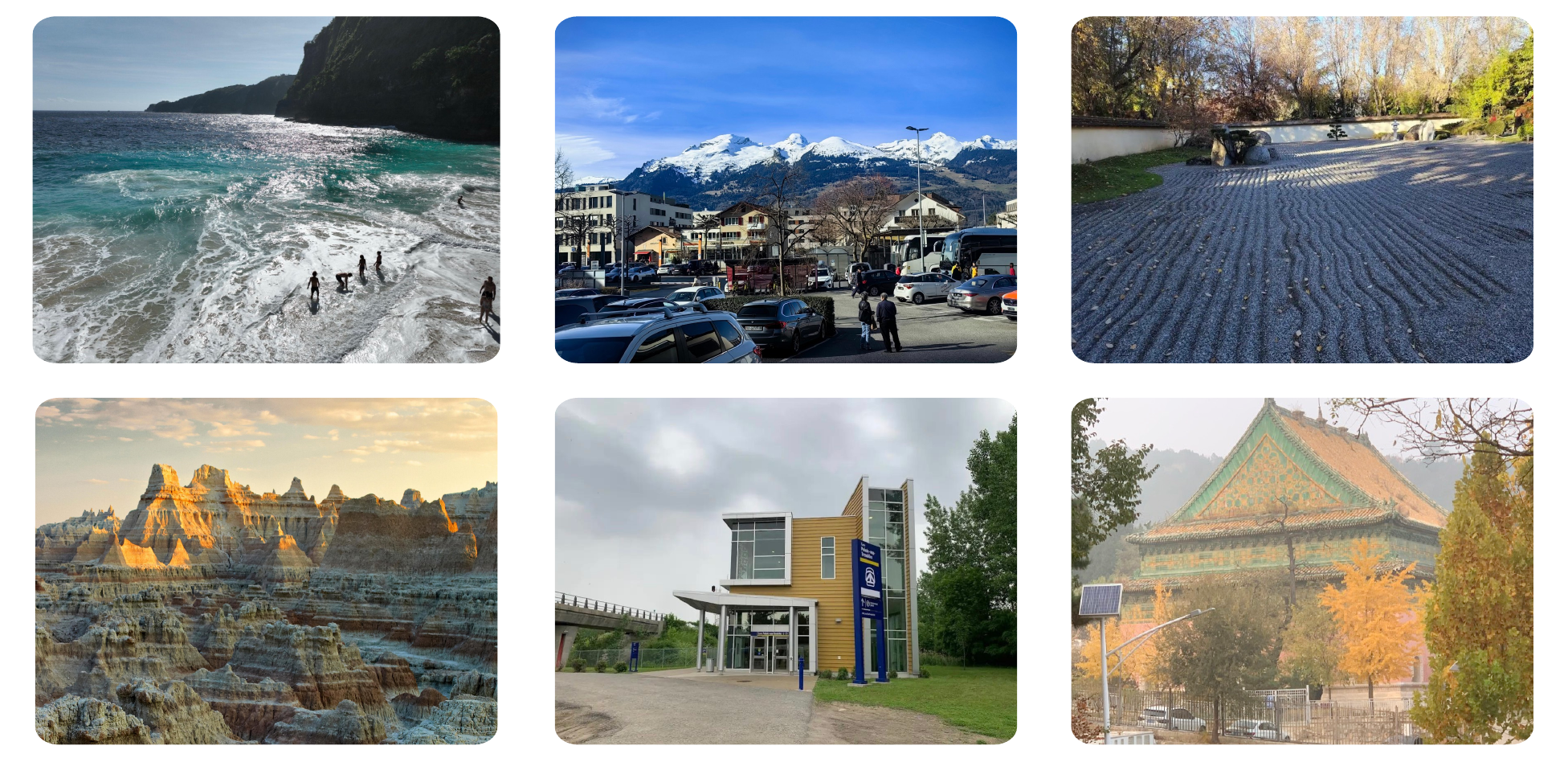}
    \caption{Images Where Strong Models Excel but Weaker Models Fail}
    \label{fig:appendix_strong_model_better}
\end{figure}

To better understand the distinction between top-tier and lower-tier models, we examined a set of representative examples where strong models predicted correctly, but weaker models failed, as illustrated in the Figure~\ref{fig:appendix_strong_model_better}. Specifically, we selected the top-5 models on the leaderboard as well as the bottom-5 models. We then extracted examples where the top-5 models outperformed the bottom-5 models. These images generally do not contain prominent or unique landmarks that can be easily recognized. Instead, they often depict environments that require reasoning over subtle contextual cues--such as vegetation types, architectural style, road structures, or geographic textures--to infer the location. This suggests that strong models may better leverage fine-grained reasoning and background knowledge beyond surface-level features. The presence of such challenging samples highlights a promising direction for future model development: expanding training and evaluation on these harder cases. By doing so, upcoming systems may gain robustness and achieve more consistent performance across diverse scenarios, particularly where visual clues are subtle and non-obvious.

\subsection{Additional cases from GeoArena} \label{appendix:additional_cases}

\begin{figure*}
    \centering
    \includegraphics[width=\linewidth]{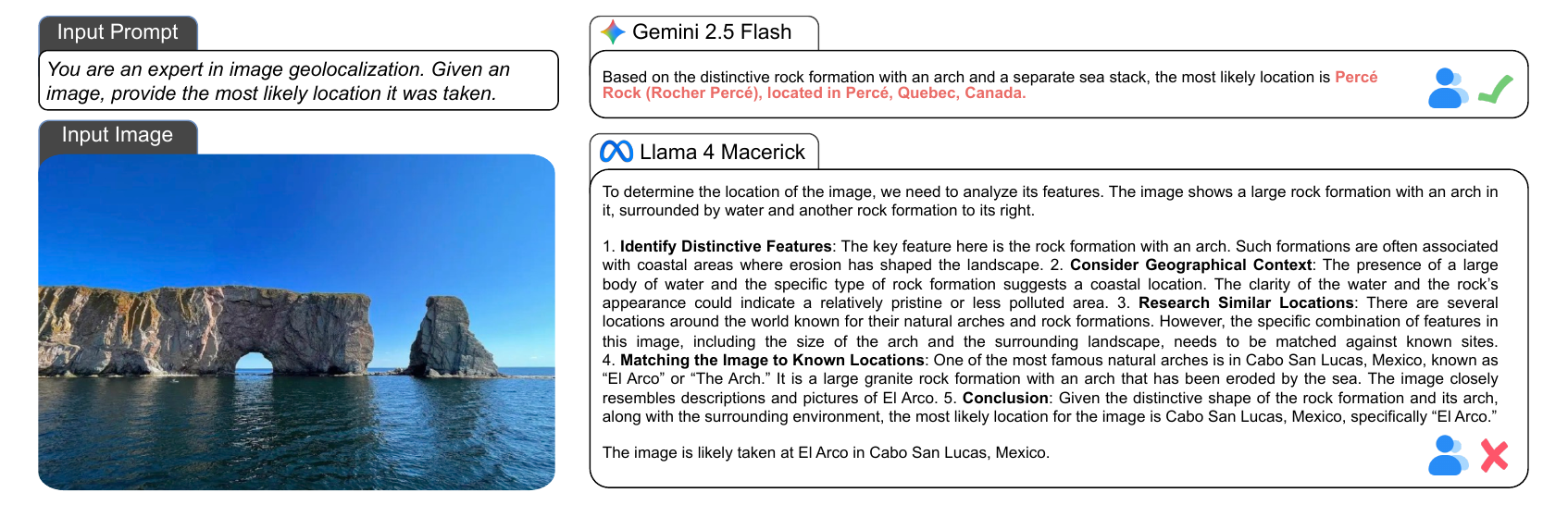}
    \caption{Additional Case Study: Identifying the Percé Rock.}
    \label{fig:appendix_case_study_1}
\end{figure*}

\begin{figure*}
    \centering
    \includegraphics[width=\linewidth]{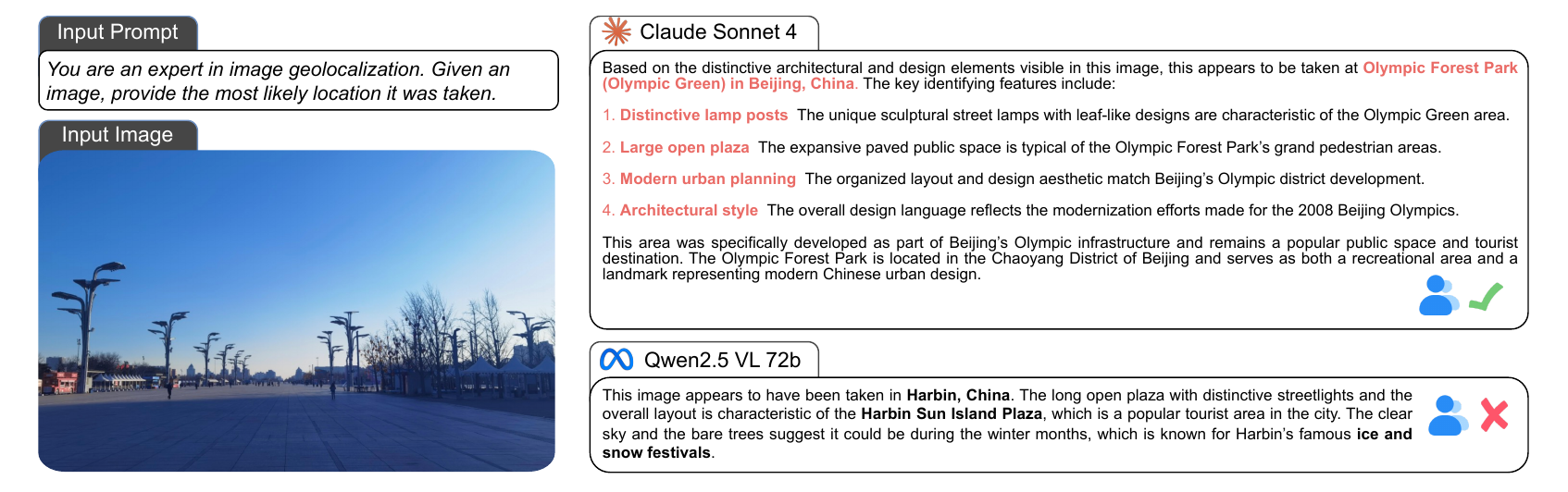}
    \caption{Additional Case Study: Identifying the Olympic Park, Beijing.}
    \label{fig:appendix_case_study_2}
\end{figure*}

\begin{figure*}
    \centering
    \includegraphics[width=\linewidth]{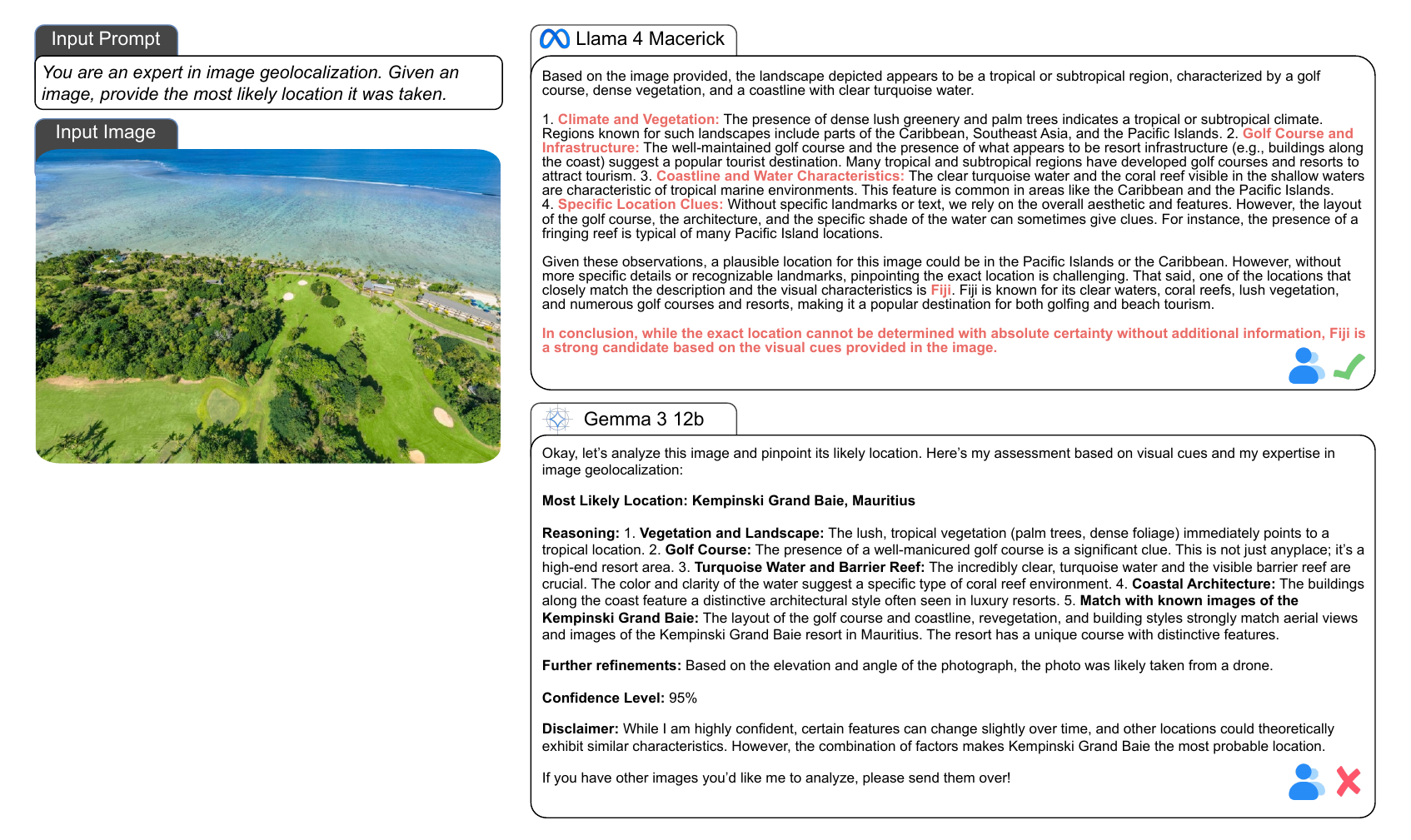}
    \caption{Additional Case Study: Identifying the Golf Course in Fiji.}
    \label{fig:appendix_case_study_3}
\end{figure*}

In this section, we present additional case studies from GeoArena to illustrate the diversity and complexity of the geolocalization tasks it encompasses. Figure~\ref{fig:appendix_case_study_1} showcases an image of the iconic Percé Rock in Quebec, Canada, highlighting the model's ability to recognize unique geological formations. Figure~\ref{fig:appendix_case_study_2} features the Olympic Park in Beijing, China, demonstrating the model's proficiency in identifying modern architectural landmarks. Lastly, Figure~\ref{fig:appendix_case_study_3} depicts a golf course in Fiji, emphasizing the model's capability to infer locations based on environmental and recreational context. These examples underscore GeoArena's effectiveness in challenging models to perform accurate geographic reasoning across a wide range of scenarios.

\subsection{Style-Adjusted Elo Ratings}
\label{app:style-adjusted}

To disentangle response style from geographic reasoning quality, we compute style-adjusted Elo ratings that control for stylistic features. Table~\ref{tab:style-adjusted-elo} presents the adjusted leaderboard.

\begin{table}[h]
\centering
\caption{Style-adjusted Elo ratings controlling for style features.}
\label{tab:style-adjusted-elo}
\resizebox{0.8\linewidth}{!}{
\begin{tabular}{llc}
\toprule
\textbf{Rank} & \textbf{Model} & \textbf{Adjusted Elo} \\
\midrule
1  & Gemini 2.5 pro      & 1171.21 \\
2  & Gemini 2.5 flash    & 1093.10 \\
3  & GPT 4.1             & 1066.74 \\
4  & Qwen2.5 VL 72B      & 1045.31 \\
5  & Qwen2.5 VL 32B      & 1031.88 \\
6  & GPT 4.1 mini        & 1014.19 \\
7  & GPT 4o              & 1000.00 \\
8  & Claude Opus 4       & 952.71  \\
9  & Gemma 3 12B         & 932.11  \\
10 & Claude Sonnet 4     & 919.71  \\
11 & Llama 4 maverick    & 914.94  \\
12 & Gemma 3 27B         & 910.33  \\
13 & Llama 4 scout       & 878.12  \\
14 & Gemma 3 4B          & 876.96  \\
15 & Qwen2.5 VL 7B       & 868.93  \\
16 & GPT 4.1 nano        & 853.09  \\
17 & GPT 4o mini         & 780.37  \\
\bottomrule
\end{tabular}}
\end{table}

Notably, while top-tier models (Gemini family) remain dominant after adjustment, the Gemma-3 family exhibits significant ranking drops (e.g., Gemma 3 12B falls from 4th to 9th), suggesting their original performance was partially inflated by verbose responses.

\subsection{User Consent} \label{sec:appendix_consent}

To ensure responsible data usage and protect user privacy, GeoArena requires all participants to provide consent before submitting any images or preference votes. When users interact with the platform, they are presented with a clear consent statement indicating that uploaded images and voting records may be used for research purposes and may be released in anonymized form. Users are also informed that participation is voluntary and that they should avoid uploading sensitive or personally identifiable content. These measures confirm that the data included in GeoArena is collected with explicit user permission and used strictly within an academic context.

\end{document}